\documentclass[journal]{IEEEtran}

\usepackage{url}
\usepackage{array,tabularx,booktabs}
\usepackage{algorithm}
\usepackage[caption=false,font=footnotesize]{subfig}
\usepackage{textcomp}
\usepackage{stfloats}
\usepackage{graphicx}
\usepackage{cite}
\usepackage{algpseudocode}
\usepackage{makecell}
\usepackage{amsmath,amssymb,amsfonts}
\usepackage[hidelinks]{hyperref}
\allowdisplaybreaks

\begin{document}

\title{Manifold-Constrained MPPI: Real-Time Sampling-Based Control for Nonlinear Equality-Constrained Robotic Systems}

\author{Seulchan~Lee,~Sanghyun~Kim%
\thanks{Seulchan Lee and Sanghyun Kim are with the School of Mechanical Engineering, Kyung Hee University, Yongin, South Korea. Sanghyun Kim is also with the Advanced Institute of Convergence Technology, Suwon, South Korea (e-mail: kim87@khu.ac.kr).}%
\thanks{Corresponding author: Sanghyun Kim.}}

\markboth{Preprint, 2026}{Lee \MakeLowercase{\textit{et al.}}: Manifold-Constrained MPPI}

\maketitle

\begin{abstract}
Sampling-based model predictive control methods, such as Model Predictive Path Integral (MPPI), offer derivative-free optimization and robustness in complex robotic systems. However, standard MPPI relies on cost-based soft penalties that cannot guarantee hard-constraint satisfaction, severely limiting its applicability to highly constrained tasks such as closed-chain manipulation. To address this, we propose Manifold-Constrained MPPI (MC-MPPI), a real-time sampling-based control framework that regulates nonlinear equality constraints while preserving the computational advantages of MPPI. The key idea is to decouple the constrained optimal control problem into latent-space planning and execution-level correction. At the planning stage, a Variational Autoencoder (VAE) learns a low-dimensional latent representation of the constraint manifold, enabling MPPI to efficiently generate near-feasible candidate trajectories without per-sample modification. Since this reference enables accurate linearization of the equality constraints, an execution-level Quadratic Programming (QP) controller resolves the residual manifold mismatch in a single solve rather than through iterative projection. Experiments on a 14-DoF closed-chain dual-arm system in both simulation and real-world settings demonstrate $100$~Hz planning and $500$~Hz execution, equality-residual regulation in static and dynamic environments, and a $95\%$ success rate over 40 randomized hardware trials. Supplementary videos and implementation details are available at \url{https://rcilab.github.io/mcmppi}.
\end{abstract}


\section{Introduction} \label{sec_1}

Robotic systems subject to strict kinematic constraints---such as the closed-chain dual-arm manipulation task \cite{hu2022practical}---require controllers that maintain constraint satisfaction in real time. Even marginal violations of these constraints can result in excessive internal forces, object damage, or complete loss of grasp stability. Enforcing kinematic constraints in robotic manipulation has been widely addressed through model predictive control with gradient-based solvers such as SQP \cite{schulman2014motion, fliege2016method}, DDP \cite{mastalli2022feasibility}, and iLQR \cite{roulet2025global}, which directly handle nonlinear constraint satisfaction. While mathematically rigorous, these approaches share common limitations: sensitivity to local minima in non-convex optimization problems and computational costs that scale poorly with the number of constraints and planning horizon. Moreover, they operate directly in the full configuration space, where the feasible set occupies a thin, nonlinear submanifold.

Model Predictive Path Integral (MPPI) \cite{williams2017model} control, a highly effective sampling-based framework for complex robotic systems, offers an alternative paradigm by replacing gradient-based optimization with a derivative-free sampling approach that is less dependent on local derivative information. Leveraging recent advancements in GPU hardware, Bhardwaj \textit{et al.} \cite{bhardwaj2022storm} introduced the STORM framework, which demonstrated that MPPI scales effectively to high-dimensional, redundant manipulators, enabling real-time trajectory optimization with thousands of parallel rollouts. Yet a fundamental limitation remains: standard MPPI relies on cost-based soft penalties to handle constraints, treating violations as additive cost terms rather than strict mathematical boundaries. This precludes any guarantee of hard-constraint satisfaction, critically limiting its applicability to tasks where physical feasibility must be maintained at all times.

To handle constraints within the MPPI framework, widely adopted approaches employ proactive safety filtering methodologies. A prominent direction utilizes Control Barrier Functions (CBFs) to delineate safe sets within the state space. For instance, to enforce safety-critical conditions such as obstacle avoidance, Ardiyanto \textit{et al.} \cite{ardiyanto2025sampling} addressed inequality constraints by formulating CBFs as soft penalties within the MPPI framework. Also, Shield-MPPI \cite{yin2023shield} introduces a discrete-time CBF-based penalty combined with a local gradient-based repair step. When equality constraints are represented only through penalties or local repair steps, however, exact satisfaction of the nonlinear manifold constraint is not guaranteed.

To incorporate constraint information directly into the sampling process, constraint-aware sampling strategies have emerged as a powerful alternative. Several recent approaches such as $\pi$-MPPI \cite{andrejev2025pi}, SCP-MPPI \cite{miura2024spline}, CSC-MPPI \cite{park2025csc}, and constrained sampling-based MPC \cite{wang2025constrained} modify sampled control sequences through projection, SVGD-based exploration, or iterative primal--dual updates. However, extending these sampling strategies to handle nonlinear equality constraints remains challenging. CSC-MPPI handles inequality constraints through iterative primal--dual correction of individual samples, whereas constrained SVGD can, in principle, accommodate both differentiable equality and inequality constraints \cite{power2024constrained}. For manifold-defined equalities, the former would require extending its per-sample updates to equality constraints, while the latter relies on tangent-space projection or equivalent manifold-aware transport, which is not incorporated into the cited SCP-MPPI formulation. Unlike inequality constraints that define feasible regions, equality constraints restrict the system to a lower-dimensional feasible manifold, making such per-sample operations particularly costly at MPPI sample counts.

To address this challenge, this paper proposes Manifold-Constrained MPPI (MC-MPPI), a novel real-time control framework that reliably regulates manifold-based equality constraints. The key idea is to decouple constraint handling into planning and execution. Rather than correcting every sampled trajectory to satisfy the constraints, MC-MPPI plans in a learned latent representation and resolves the constraint mismatch only once for the selected solution at execution. At the planning stage, inspired by recent learning-based approaches to constraint-manifold sampling for robot motion planning~\cite{lembono2021learning, park2024constrained}, a Variational Autoencoder (VAE)~\cite{kingma2019introduction} is employed to learn a continuous, lower-dimensional latent representation of the constraint manifold. Instead of sampling in the high-dimensional full configuration space---where nearly all samples would violate the zero-measure equality constraints---the framework samples directly in this structured latent space. This enables MPPI to efficiently generate thousands of candidate trajectories that are structurally near-feasible, while avoiding the prohibitive cost of modifying every individual sample. However, because a learned generative model inevitably introduces approximation errors, a manifold mismatch remains. To address this, at the execution stage, a QP-based optimal controller~\cite{kim2019continuous} is employed exclusively on the selected solution to resolve the residual manifold mismatch.

Therefore, the primary contributions of this study are summarized as follows:
\begin{itemize}
    \item \textbf{A Manifold-Constrained MPPI Framework for Nonlinear Equality Constraints:}
    We propose MC-MPPI, a real-time sampling-based control framework that reliably regulates manifold-based equality constraints. By decoupling the constrained optimal control problem into latent-space planning and execution, the framework preserves the derivative-free, parallelizable advantages of MPPI while maintaining near-manifold operation.

    \item \textbf{Computationally Efficient Resolution of Manifold Mismatch via Single-Step QP:}
    To resolve the manifold mismatch inherent in learned generative models, we introduce a computationally efficient two-stage architecture. Because the latent-space planning provides a structurally near-feasible reference, the nonlinear equality constraints can be locally linearized. This enables the execution-level QP controller to regulate the residual with a single solve per cycle rather than iterative manifold projection, maintaining stable real-time operation at high control frequency.

    \item \textbf{Experimental Validation in Simulation and on a Real Closed-Chain Dual-Arm Setup:}
    The proposed framework is validated through both simulation and hardware experiments. The results demonstrate $100$~Hz planning and $500$~Hz execution on the 14-DoF closed-chain dual-arm system together with equality-residual regulation in static and dynamic obstacle scenarios.
\end{itemize}
Accordingly, we propose a unified control framework that combines manifold-aware latent-space planning with high-rate residual regulation. The proposed architecture extends derivative-free MPPI to nonlinear equality-constrained robotic systems without requiring per-sample projection and employs constant innovation sampling in the learned latent space to generate smooth and directionally consistent exploration.

\section{Review of MPPI}
\label{sec_2}

For readability, we adopt the following notation conventions throughout this paper: $\tilde{(\cdot)}$, $\hat{(\cdot)}^{*}$, and $(\cdot)^*$ denote perturbed, near-feasible, and optimal quantities, respectively.

MPPI is a sampling-based optimal control framework that replaces gradient-based optimization with importance sampling over the control space. Consider a discrete-time stochastic nonlinear system:
\begin{equation}
    \boldsymbol{x}_{t+1} = f(\boldsymbol{x}_t, \tilde{\boldsymbol{u}}_t),
\label{eq_1}
\end{equation}
where $\boldsymbol{x}_t \in \mathbb{R}^{n_s}$ is the state and $\tilde{\boldsymbol{u}}_t \in \mathbb{R}^{n_u}$ is the perturbed control input at time step $t$. The controller maintains a control sequence $\boldsymbol{U} = [\boldsymbol{u}_0, \ldots, \boldsymbol{u}_{T-1}]$ over a finite horizon $T$, where each perturbed control is generated as:
\begin{equation}
    \tilde{\boldsymbol{u}}_t = \boldsymbol{u}_t + \delta\boldsymbol{u}_t, \quad \delta\boldsymbol{u}_t \sim \mathcal{N}(\boldsymbol{0}, \boldsymbol{\Sigma}),
\label{eq_2}
\end{equation}
where $\boldsymbol{\Sigma}$ is the exploration covariance matrix. Here, $\delta\boldsymbol{u}_t$ is a Gaussian
perturbation used for stochastic optimization in MPPI. For each sampled control trajectory $\tilde{\boldsymbol{U}} = [\tilde{\boldsymbol{u}}_0, \ldots, \tilde{\boldsymbol{u}}_{T-1}]$, the state sequence, $\boldsymbol{X} = [\boldsymbol{x}_0, \ldots, \boldsymbol{x}_T]$, is obtained by forward propagation through \eqref{eq_1}, and its performance is evaluated by the trajectory cost:
\begin{equation}
    S(\boldsymbol{X}) = \phi(\boldsymbol{x}_T) + \sum_{t=0}^{T-1} c(\boldsymbol{x}_t),
\label{eq_3}
\end{equation}
 where $\phi(\cdot)$ and $c(\cdot)$ denote the user-defined, task-dependent terminal and stage costs, respectively.

\begin{figure*}[t]
\centering
\includegraphics[width=1.0\linewidth]{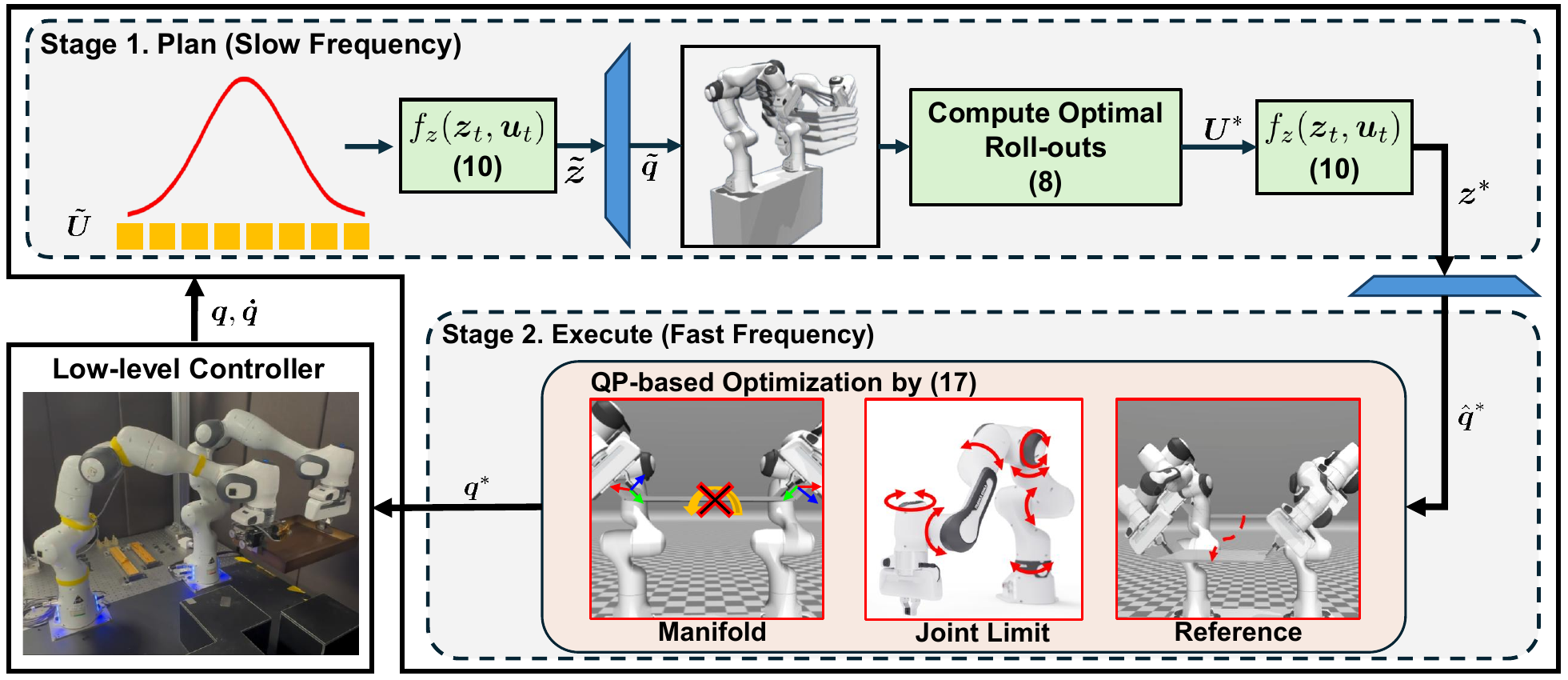}
\caption{Overall architecture of the MC-MPPI framework. The upper-level planner performs MPPI in a VAE-learned latent space to efficiently generate near-feasible candidate trajectories under manifold-based equality constraints. The decoded nominal solution is then corrected by a single-step QP that regulates the equality residual, and the resulting reference is tracked by an optimization-based low-level controller. In our experiments with the dual-arm system, the slow-frequency planner operates at 100 Hz, while the fast-frequency executor runs at 500 Hz.}
\label{fig_1}
\end{figure*}

The objective is to find the optimal control sequence $\boldsymbol{U}^*$ that minimizes $\mathbb{E}[S(\boldsymbol{X})]$, but a direct solution is intractable. By applying the free energy formulation and Jensen's inequality, the stochastic optimal control problem can be reformulated as minimizing the KL divergence between the optimal trajectory distribution $\mathbb{Q}^*$ and the proposal distribution $\mathbb{Q}$~\cite{williams2017model}:
\begin{equation}
    \boldsymbol{U}^* \simeq \arg\min_{\boldsymbol{U}} \; D_{\mathrm{KL}}\!\left(\mathbb{Q}^* \,\|\, \mathbb{Q}\right),
\label{eq_4}
\end{equation}
whose solution yields:
\begin{align}
    \boldsymbol{U}^* &= \arg\min_{\boldsymbol{U}} \;
    \mathbb{E}_{\mathbb{Q}^*}\!\left[\frac{1}{2}
    \sum_{t=0}^{T-1}(\tilde{\boldsymbol{u}}_t - \boldsymbol{u}_t)^T
    \boldsymbol{\Sigma}^{-1}(\tilde{\boldsymbol{u}}_t -
    \boldsymbol{u}_t)\right] \nonumber \\
    &= \mathbb{E}_{\mathbb{Q}^*}\!\left[
    \tilde{\boldsymbol{U}}\right],
    \label{eq_5}
\end{align}
stating that the optimal control sequence is the expectation under $\mathbb{Q}^*$. Here, $\boldsymbol{U}$ is the optimization variable updated by MPPI, while $\delta\boldsymbol{u}_t$ is a sampled perturbation and is not optimized separately.
Since sampling directly from $\mathbb{Q}^*$ is intractable, importance sampling is applied to express the expectation under the proposal distribution $\mathbb{Q}$. The resulting importance weights are approximated via $K$ Monte Carlo samples. For each sample $k \in \{1, \ldots, K\}$, a perturbed control sequence is generated by instantiating (\ref{eq_2}) as:
\begin{equation}
    \tilde{\boldsymbol{u}}^{(k)}_t = \boldsymbol{u}_t + \delta\boldsymbol{u}_t^{(k)}, \quad \delta\boldsymbol{u}_t^{(k)} \sim \mathcal{N}(\boldsymbol{0}, \boldsymbol{\Sigma}),
\label{eq_6}
\end{equation}
and rolled out through \eqref{eq_1} to obtain the trajectory cost $S^{(k)}$. Each sample is assigned the importance weight:
\begin{equation}
    w^{(k)} = \frac{\exp\!\left(-\frac{1}{\lambda} S^{(k)}\right)}{\sum_{j=1}^{K}\exp\!\left(-\frac{1}{\lambda}S^{(j)}\right)},
\label{eq_7}
\end{equation}
where $\lambda > 0$ is the temperature parameter. The optimal control sequence is computed as the importance-weighted average:
\begin{equation}
    \boldsymbol{U}^{*} \leftarrow \sum_{k=1}^{K} w^{(k)} \tilde{\boldsymbol{U}}^{(k)}.
\label{eq_8}
\end{equation}
The first action $\boldsymbol{u}_0$ is applied to the system, the horizon is shifted forward, and the process repeats in a receding-horizon fashion.

While standard MPPI efficiently handles non-convex cost landscapes through massively parallel sampling, it handles constraints only through cost penalties and therefore does not guarantee their satisfaction. This motivates the manifold-constrained formulation proposed in this work.

\section{MC-MPPI Framework}
\label{sec_3}

This section presents the two-stage MC-MPPI architecture shown in
Fig.~\ref{fig_1}, consisting of the planning stage in
Subsection~\mbox{\ref{sec_31}} and the execution stage in
Subsection~\mbox{\ref{sec_32}}.

\subsection{Planning Stage}
\label{sec_31}
\subsubsection{Manifold Learning via VAE}
\label{sec_311}

Consider an $n$-dimensional robotic system subject to $l$ kinematic equality constraints defined by $\boldsymbol{h}: \mathbb{R}^n \to \mathbb{R}^l$ with $\boldsymbol{h}(\boldsymbol{q})=\boldsymbol{0}$, which induce a nonlinear constraint manifold
\begin{equation}
    \mathcal{M} = \left\{ \boldsymbol{q} \in \mathbb{R}^n \mid \boldsymbol{h}(\boldsymbol{q}) = \boldsymbol{0} \right\},
\label{eq_9}
\end{equation}
which, assuming that $\boldsymbol{J}_h$ has full row rank $l$ on the operating portion of $\mathcal{M}$, has intrinsic dimension $m=n-l$. Planning directly on $\mathcal{M}$ is difficult because it is a lower-dimensional set embedded in $\mathbb{R}^n$, so naive sampling in the ambient space yields feasible configurations with probability zero. To address this, a VAE is employed to learn a continuous, low-dimensional latent space that captures the structure of $\mathcal{M}$. The pre-trained decoder $\psi_\theta: \mathbb{R}^m \to \mathbb{R}^n$ then provides a nonlinear map that approximately parameterizes $\mathcal{M}$, enabling trajectory optimization to be performed in the unconstrained latent space while producing near-feasible configurations. The VAE prior regularizes latent codes toward a Gaussian distribution, making the learned space compatible with MPPI's Gaussian exploration. This regularization does not guarantee exact feasibility; the execution stage regulates the residual mismatch. During online planning, the decoder maps latent states to joint configurations, while the encoder is used only for training and initialization.

\subsubsection{Latent Space Control Formulation}
To perform manifold-constrained sampling within the MPPI framework, candidate trajectories must be generated in a space where feasible samples can be obtained efficiently. To this end, the proposed framework executes latent motion within the learned latent space $\mathbb{R}^m$ of the VAE, rather than sampling directly in the ambient configuration space $\mathbb{R}^n$, where feasible configurations form a zero-measure subset under any continuous distribution.

To realize this latent motion, we consider stochastic latent-space
dynamics analogous to \mbox{\eqref{eq_1}}.
Here, $\boldsymbol{z}_t$ plays the role of the state
$\boldsymbol{x}_t$ in Section~\ref{sec_2}.
The latent control input $\boldsymbol{u}_t\in\mathbb{R}^{m}$ is
interpreted as the latent-space velocity $\dot{\boldsymbol{z}}_t$.
Accordingly, the latent evolution is modeled by the first-order Euler
integration
\begin{equation}
\boldsymbol{z}_{t+1}
=
f_z(\boldsymbol{z}_t,\tilde{\boldsymbol{u}}_t)
=
\boldsymbol{z}_t+\tilde{\boldsymbol{u}}_t\Delta t,
\label{eq_10}
\end{equation}
The corresponding physical joint configuration is obtained by decoding the latent state,
$\boldsymbol{q}_t=\psi_\theta(\boldsymbol{z}_t)$. Since the learned manifold is defined in configuration space,
the decoder generates a kinematic joint reference rather than a torque
command, which is subsequently tracked by the low-level controller. As in the standard MPPI formulation in \eqref{eq_2}, the perturbed
latent control input is given by
\begin{equation}
\tilde{\boldsymbol{u}}_t
=
\boldsymbol{u}_t+\delta\boldsymbol{u}_t,
\qquad
\delta\boldsymbol{u}_t
\sim
\mathcal{N}(\boldsymbol{0},\boldsymbol{\Sigma}).
\label{eq_11}
\end{equation}

Based on this formulation, the stochastic optimal control problem is defined over a horizon $T$ with latent control sequence $\boldsymbol{U}=[\boldsymbol{u}_0,\boldsymbol{u}_1,\dots,\boldsymbol{u}_{T-1}]^T \in \mathbb{R}^{T \times m}$ as
\begin{equation}
\begin{aligned}
    \min_{\boldsymbol{U}} \; & J = \mathbb{E}\left[
    \phi(\boldsymbol{z}_{T}) + \sum_{t=0}^{T-1}
    \left( c(\boldsymbol{z}_t) + \frac{1}{2}\boldsymbol{u}_t^T \boldsymbol{R}\boldsymbol{u}_t \right)
    \right], \\
    \text{s.t.} \quad
    & \boldsymbol{z}_{t+1} = \boldsymbol{z}_t + \tilde{\boldsymbol{u}}_t \Delta t, \\
    & \tilde{\boldsymbol{u}}_t = {\boldsymbol{u}}_t + \delta\boldsymbol{u}_t,\quad
      \delta\boldsymbol{u}_t \sim \mathcal{N}(\boldsymbol{0}, \boldsymbol{\Sigma}), \\
    & \boldsymbol{z}_0 = \boldsymbol{z}_{\mathrm{init}},
\end{aligned}
\label{eq_12}
\end{equation}
where $c(\boldsymbol{z}_t)$ and $\phi(\boldsymbol{z}_{T})$ denote the stage and terminal costs, respectively, and $\boldsymbol{R}\succ0$ weights the standard quadratic path-integral control cost $\tfrac{1}{2}\boldsymbol{u}_t^{T}\boldsymbol{R}\boldsymbol{u}_t$.

\subsubsection{MPPI in Latent Space}
\label{sec_313}
Since the optimization problem in \mbox{\eqref{eq_12}} is formulated in the latent space, costs defined in the joint and operational spaces---such as end-effector tracking error, collision avoidance, and joint limits---cannot be evaluated directly from latent samples $\tilde{\boldsymbol{z}}$. Therefore, each predicted latent state along a sampled trajectory is decoded into the joint space using the pre-trained decoder:
\begin{equation}
    \tilde{\boldsymbol{q}}^{(k)}_{t+1} = \psi_{\theta}(\tilde{\boldsymbol{z}}^{(k)}_{t+1}), \quad \forall t=0, \dots, T-1
\label{eq_13}
\end{equation}
where $\psi_{\theta}$ denotes the pre-trained decoder described in Subsection~\mbox{\ref{sec_311}}. The decoded trajectory $\tilde{\boldsymbol{q}}^{(k)}$ is then used to evaluate the trajectory cost $S^{(k)}$ in the joint and operational spaces, enabling meaningful cost comparison across all $K$ sampled trajectories.

However, the quality of the decoded joint trajectory $\tilde{\boldsymbol{q}}^{(k)}$ is sensitive to the temporal continuity of the latent samples $\tilde{\boldsymbol{z}}^{(k)}$. To promote temporally consistent latent exploration while maintaining computational efficiency, we adopt the single-instance sampling strategy proposed in \cite{kim2025single}, which maintains a constant control innovation $\delta\boldsymbol{u}^{(k)}$ over the predictive horizon:
\begin{equation}
    \tilde{\boldsymbol{u}}_{t}^{(k)} = \boldsymbol{u}_t + \delta\boldsymbol{u}^{(k)}, \quad \delta\boldsymbol{u}^{(k)} \sim \mathcal{N}(\boldsymbol{0}, \boldsymbol{\Sigma}),
\label{eq_14}
\end{equation}
where $\delta\boldsymbol{u}^{(k)} \in \mathbb{R}^{m}$ is a trajectory-specific noise vector sampled once and applied uniformly over the entire prediction horizon.

For each sample $k$, the resulting latent trajectory $\{\tilde{\boldsymbol{z}}_{t}^{(k)}\}_{t=1}^{T}$ is decoded via pre-trained decoder $\psi_{\theta}$ to obtain the joint trajectory $\{\tilde{\boldsymbol{q}}_{t}^{(k)}\}_{t=1}^{T}$, from which the trajectory cost $S^{(k)}$ is evaluated in the operational space. Once all trajectory costs $\{S^{(k)}\}_{k=1}^{K}$ are evaluated, the importance weights $\{w^{(k)}\}_{k=1}^{K}$ are computed as in \mbox{\eqref{eq_7}}, and the optimal latent control sequence $\boldsymbol{U}^{*}$ is updated as the importance-weighted average of the perturbed sequences, analogously to \mbox{\eqref{eq_8}}. Here, $\boldsymbol{U}^{*} = [\boldsymbol{u}_0^{*}, \boldsymbol{u}_1^{*}, \dots, \boldsymbol{u}_{T-1}^{*}]^T$ denotes the updated optimal latent control sequence corresponding to the sequence $\boldsymbol{U}$. The optimal latent state is then updated via first-order Euler integration, consistent with the latent dynamics defined in \mbox{\eqref{eq_10}}:
\begin{equation}
    \boldsymbol{z}^{*} = \boldsymbol{z}_c + \boldsymbol{u}^{*}_{0} \Delta t,
\label{eq_15}
\end{equation}
where $\boldsymbol{z}_c$ denotes the current latent state, propagated from the previous optimal solution. Finally, the optimal joint configuration is recovered by decoding $\boldsymbol{z}^{*}$ as
\begin{equation}
    \hat{\boldsymbol{q}}^{*} = \psi_{\theta}(\boldsymbol{z}^{*}),
\label{eq_16}
\end{equation}
where $\hat{\boldsymbol{q}}^{*}$ is the near-feasible configuration produced by the planning stage and passed to the execution stage as a reference.

\subsection{Execution Stage}
\label{sec_32}
Although the planning stage yields a near-feasible reference $\hat{\boldsymbol{q}}^{*}$, it does not lie exactly on $\mathcal{M}$ due to approximation errors of the learned decoder, a phenomenon referred to as \textit{manifold mismatch}. The execution stage therefore regulates this residual before passing the reference to the low-level controller. %

To regulate the residual constraint violation while preserving task relevance, we formulate a single-step QP that drives the next-state configuration $\boldsymbol{q}^{*}$ toward $\mathcal{M}$. Linearizing the closure constraint $\boldsymbol{h}(\boldsymbol{q}) = \boldsymbol{0}$ to first order around the current configuration $\boldsymbol{q}_c$ yields the following formulation:
\begin{equation}
\begin{aligned}
    \dot{\boldsymbol{q}}^{*}, \boldsymbol{q}^{*} = \arg\min_{\dot{\boldsymbol{q}}, \boldsymbol{q}} \; & \| \dot{\boldsymbol{q}} - \dot{\boldsymbol{q}}_{\text{ref}} \|^2 + w_{\text{task}} \| \boldsymbol{J}_{\text{task}} \dot{\boldsymbol{q}} - \dot{\boldsymbol{x}}_{\text{task}}^{*} \|^2 \\
    \text{s.t.} \; & \boldsymbol{q} = \boldsymbol{q}_c + \dot{\boldsymbol{q}} \Delta t_e, \\
    & \boldsymbol{J}_h \dot{\boldsymbol{q}} = -\alpha \boldsymbol{h}(\boldsymbol{q}_c), \\
    & \dot{\boldsymbol{q}}_{\min} \leq \dot{\boldsymbol{q}} \leq \dot{\boldsymbol{q}}_{\max},
\end{aligned}
\label{eq_17}
\end{equation}
where $\boldsymbol{q}_c$ denotes the current joint configuration, $\boldsymbol{J}_h = \partial \boldsymbol{h} / \partial \boldsymbol{q} \big|_{\boldsymbol{q}_c}$ is the constraint Jacobian evaluated at $\boldsymbol{q}_c$, the reference velocity $\dot{\boldsymbol{q}}_{\text{ref}} = (\hat{\boldsymbol{q}}^* - \boldsymbol{q}_c) / \Delta t_e$ tracks the MC-MPPI solution, and the desired task velocity $\dot{\boldsymbol{x}}_{\text{task}}^{*} = -K_{p,\text{task}} \boldsymbol{e}_{\text{task}}(\boldsymbol{q}_c)$ regulates the operational-space error $\boldsymbol{e}_{\text{task}}$ (e.g., the pose error of the tray center or the left end-effector) through the task Jacobian $\boldsymbol{J}_{\text{task}}$, weighted by $w_{\text{task}}$. Here, $\Delta t_e$ is the execution period, which is shorter than the planning step $\Delta t$. The first equality relates the next configuration $\boldsymbol{q}^{*}$ to $\boldsymbol{q}_c$ via Euler integration, while the second drives the linearized constraint residual $\boldsymbol{h}(\boldsymbol{q}_c)$ toward zero at rate $\alpha$. The velocity bounds $\dot{\boldsymbol{q}}_{\min} = (\boldsymbol{q}_{\text{lb}} - \boldsymbol{q}_c) / \Delta t_e$ and $\dot{\boldsymbol{q}}_{\max} = (\boldsymbol{q}_{\text{ub}} - \boldsymbol{q}_c) / \Delta t_e$ enforce the joint position limits as \emph{hard} box constraints, while obstacle avoidance is left to the soft MPPI cost.

\begin{algorithm}[t]
\caption{Planning and Execution of MC-MPPI}
\label{alg_1}
{
\begin{algorithmic}[1]

\Require
\Statex \hspace{1em} $\psi_{\theta}$: Pre-trained decoder
\Statex \hspace{1em} $\boldsymbol{z}_c$: Current latent state
\Statex \hspace{1em} $\boldsymbol{q}_c$: Current joint configuration
\Statex \hspace{1em} $\boldsymbol{U}=[\boldsymbol{u}_0,\ldots,\boldsymbol{u}_{T-1}]$: Nominal control sequence
\Statex \hspace{1em} $\boldsymbol{\Sigma}$: Noise covariance
\Statex \hspace{1em} $K,T$: Number of samples and horizon

\Statex \emph{\textbf{Planning stage (100~Hz):}}
\For{$k=1,\ldots,K$} \Comment{Parallel sampling}
    \State $\delta\boldsymbol{u}^{(k)}
    \sim\mathcal{N}(\boldsymbol{0},\boldsymbol{\Sigma})$
    \Comment{Single-instance sampling}
    \State $\tilde{\boldsymbol{z}}_0^{(k)}
    \leftarrow \boldsymbol{z}_c$
    \For{$t=0,\ldots,T-1$}
        \State $\tilde{\boldsymbol{u}}_t^{(k)}
        \leftarrow
        \boldsymbol{u}_t+\delta\boldsymbol{u}^{(k)}$
        \State $\tilde{\boldsymbol{z}}_{t+1}^{(k)}
        \leftarrow
        \tilde{\boldsymbol{z}}_t^{(k)}
        +\tilde{\boldsymbol{u}}_t^{(k)}\Delta t$
        \State $\tilde{\boldsymbol{q}}_{t+1}^{(k)}
        \leftarrow
        \psi_{\theta}(\tilde{\boldsymbol{z}}_{t+1}^{(k)})$
        \Comment{\eqref{eq_13}}
    \EndFor
    \State Evaluate trajectory cost $S^{(k)}$ using
    $\{\tilde{\boldsymbol{z}}_t^{(k)},
    \tilde{\boldsymbol{q}}_t^{(k)}\}_{t=1}^{T}$
\EndFor

\State Compute importance weights
$\{w^{(k)}\}_{k=1}^{K}$ via \eqref{eq_7}

\State $\boldsymbol{U}^{*}
\leftarrow
\sum_{k=1}^{K}w^{(k)}
\tilde{\boldsymbol{U}}^{(k)}$
\Comment{\eqref{eq_8}}

\State $\boldsymbol{U}
\leftarrow
[\boldsymbol{u}^{*}_{1},\ldots,
\boldsymbol{u}^{*}_{T-1},\boldsymbol{0}]$
\Comment{Horizon shift}

\State $\boldsymbol{z}^{*}
\leftarrow
\boldsymbol{z}_c+\boldsymbol{u}^{*}_{0}\Delta t$
\Comment{\eqref{eq_15}}

\State $\hat{\boldsymbol{q}}^{*}
\leftarrow
\psi_{\theta}(\boldsymbol{z}^{*})$
\Comment{\eqref{eq_16}}

\Statex \emph{\textbf{Execution stage (500~Hz):}}

\State $\boldsymbol{h}_c
\leftarrow
\boldsymbol{h}(\boldsymbol{q}_c),\quad
\boldsymbol{J}_h
\leftarrow
\boldsymbol{J}_h(\boldsymbol{q}_c)$

\State Solve QP~\eqref{eq_17} for
$\dot{\boldsymbol{q}}^{*}$

\If{QP feasible}
    \State $\boldsymbol{q}^{*}
    \leftarrow
    \boldsymbol{q}_c+
    \dot{\boldsymbol{q}}^{*}\Delta t_e$
\Else
    \State $\boldsymbol{q}^{*}
    \leftarrow
    \boldsymbol{q}^{*}_{\mathrm{prev}}$
    \Comment{Fallback}
\EndIf

\State \Return $\boldsymbol{q}^{*}$ to the low-level controller

\end{algorithmic}
}
\end{algorithm}

The velocity-level equality in~\eqref{eq_17} is imposed directly as an equality
constraint within the execution QP rather than as an additive penalty.
Because the original constraint
$\boldsymbol{h}(\boldsymbol{q})=\boldsymbol{0}$ is nonlinear and the
computed velocity is applied over a finite execution step, a residual
may remain after a single linearized correction.
The residual is therefore re-evaluated and corrected at every
execution cycle at $500$~Hz. Accordingly, the execution layer provides high-rate regulation of the
nonlinear constraint residual, rather than exact one-step projection
onto the constraint manifold.
The velocity bounds enforce the joint position limits as hard box
constraints, whereas obstacle avoidance is handled through the soft
MPPI rollout cost.

Although $\hat{\boldsymbol{q}}^*$ is computed from near-feasible samples of the planning stage, the QP-based optimization in \mbox{\eqref{eq_17}} may not preserve all of its task-relevant features, $\boldsymbol{e}_{task}$, because the optimization direction toward $\mathcal{M}$ is governed by the constraint Jacobian $\boldsymbol{J}_h$ rather than the task objective. To address this, the task-tracking term explicitly biases the optimization toward the task-relevant region of $\mathcal{M}$, ensuring that the resulting configuration $\boldsymbol{q}^{*}$ aligns with the actual task target.

The resulting $\boldsymbol{q}^{*}$ is then passed to a low-level controller for execution on the robotic system.\footnote{Note that the proposed framework is agnostic to the choice of low-level controller; in this work, a position controller is used in our closed-chain dual-arm implementation.} In the rare event that the QP becomes infeasible, the framework maintains the previous optimal configuration ($\boldsymbol{q}^{*} = \boldsymbol{q}^{*}_{\text{prev}}$) to preserve command continuity.  This fallback does not by itself guarantee safety in dynamic environments.\footnote{It is worth noting that throughout all simulations and hardware experiments conducted in this study, the single-step QP consistently yielded a feasible solution, and this fallback condition was never triggered.} The complete MC-MPPI procedure is summarized in Algorithm~\ref{alg_1}.

\section{Experiments}
\label{sec_4}

This section validates the effectiveness of the proposed MC-MPPI framework through both simulation and real-world experiments. To this end, three main experiments are conducted. First, the framework is evaluated to validate its capability for handling constraints, demonstrating the essential role of the QP-based execution stage in resolving the manifold mismatch inherent in the learned VAE decoder. Second, the framework's capability for static obstacle avoidance is evaluated, accompanied by an analysis of constant-innovation latent rollouts to characterize their effect on exploration efficiency and trajectory smoothness within the nonlinear latent space. Finally, the reactiveness of MC-MPPI is assessed in an environment with dynamic obstacles. This final experiment is conducted on a real-world closed-chain dual-arm system, demonstrating both reactive capability and practical applicability.

\subsection{Experiments Overview}
\label{sec_41}

To validate the proposed MC-MPPI framework on manifold-based equality constraints, a closed-chain dual-arm manipulation task is adopted as a representative benchmark. The system consists of two Panda manipulators (\textit{Franka Emika Co}.) jointly grasping a flat tray, with configuration $\boldsymbol{q} \in \mathbb{R}^{14}$ subject to the equality constraint $\boldsymbol{h}(\boldsymbol{q}) = [\boldsymbol{h}_{\mathrm{cc}}(\boldsymbol{q})^{T},\, \boldsymbol{h}_{\mathrm{flat}}(\boldsymbol{q})^{T}]^{T} \in \mathbb{R}^{8}$, comprising the following two terms.

The first is a 6-D closed-chain term that enforces the relative $SE(3)$ pose between the two end-effectors:
\begin{equation}
\boldsymbol{h}_{\mathrm{cc}}(\boldsymbol{q}) = \log\!\left( \left({}^{w}\boldsymbol{T}_{l} \, {}^{l}\boldsymbol{T}_{r}^{*}\right)^{-1} \, {}^{w}\boldsymbol{T}_{r} \right) \in \mathbb{R}^{6},
\end{equation}
where ${}^{w}\boldsymbol{T}_{l},\, {}^{w}\boldsymbol{T}_{r} \in SE(3)$ denote the left and right end-effector poses in the world frame, and ${}^{l}\boldsymbol{T}_{r}^{*}$ is the fixed relative transform required to maintain the bimanual grasp. The second is a 2-D tray flatness term enforcing zero roll and pitch about the tray center:
\begin{equation}
\boldsymbol{h}_{\mathrm{flat}}(\boldsymbol{q}) = [\phi_{\mathrm{tray}},\, \theta_{\mathrm{tray}}]^{T} = \boldsymbol{0} \in \mathbb{R}^{2}.
\end{equation}
The resulting constraint manifold $\mathcal{M} = \{\boldsymbol{q} \in \mathbb{R}^{14} \mid \boldsymbol{h}(\boldsymbol{q}) = \boldsymbol{0}\}$ has an intrinsic dimension of $n - l = 6$, and the VAE is accordingly trained with a latent dimension of $m = 6$. The $\beta$-VAE has a symmetric $14$--$(8{\times}256)$--$6$ MLP architecture with GELU activations. It is trained on $80000$ feasible configurations using AdamW with a learning rate of $10^{-3}$. The loss combines an $L_1$ reconstruction term with weight $5.0$ and a KL term whose weight is annealed from $0.001$ to $0.02$.

The experiments in Subsections \mbox{\ref{sec_42}} and \mbox{\ref{sec_43}} are conducted in MuJoCo~\mbox{\cite{todorov2012mujoco}}, a high-fidelity physics simulator with full rigid-body dynamics and contact resolution. The MC-MPPI planner uses $K = 200$ samples, a prediction horizon of $T = 30$, a temperature of $\lambda=0.005$, and a planning step of $\Delta t = 0.01$~s; the execution period is $\Delta t_e = 0.002$~s. Success is defined as preserving the bimanual grasp while reducing the position and orientation errors below $0.01$~m and $0.01$~rad, respectively. Constraint violation is measured by $\|\boldsymbol{h}(\boldsymbol{q})\|$, which combines position~(m) and orientation~(rad) residuals. All computations are performed on a computer with an Intel Core i5-13400F CPU, 16~GB RAM, and an NVIDIA GeForce RTX 4060 Ti GPU.

During the MPPI planning stage, the trajectory cost $S$ is evaluated as a weighted sum of five principal components:
\begin{equation}
    S = S_{\text{track}} + S_{\text{coll}} + S_{\text{reg}} + S_{\text{limit}} + S_{\text{neutral}},
\end{equation}
where $S_{\text{track}}$ penalizes end-effector tracking error, $S_{\text{coll}}$ penalizes self-collision and environmental collision, $S_{\text{reg}}$ is the quadratic latent-control penalty $\boldsymbol{u}^{T}\boldsymbol{R}\boldsymbol{u}$, $S_{\text{limit}}$ penalizes joint-position and latent-velocity limit violations, and $S_{\text{neutral}}$ discourages drift from a neutral joint posture. The detailed cost terms, hyperparameters, and VAE settings are provided on the project website at \mbox{\url{https://rcilab.github.io/mcmppi}}.

\begin{figure}[!t]
\centering

\subfloat[]{%
    \includegraphics[width=\linewidth]{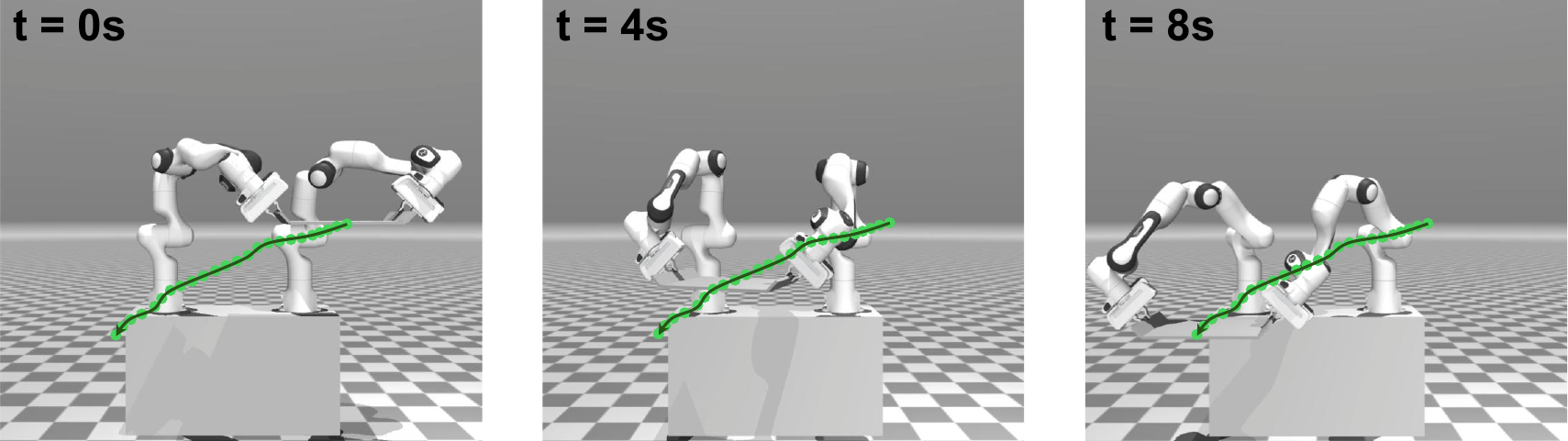}%
    \label{fig_2a}%
}\\[-0.4em]
\subfloat[]{%
    \includegraphics[width=0.455\linewidth]{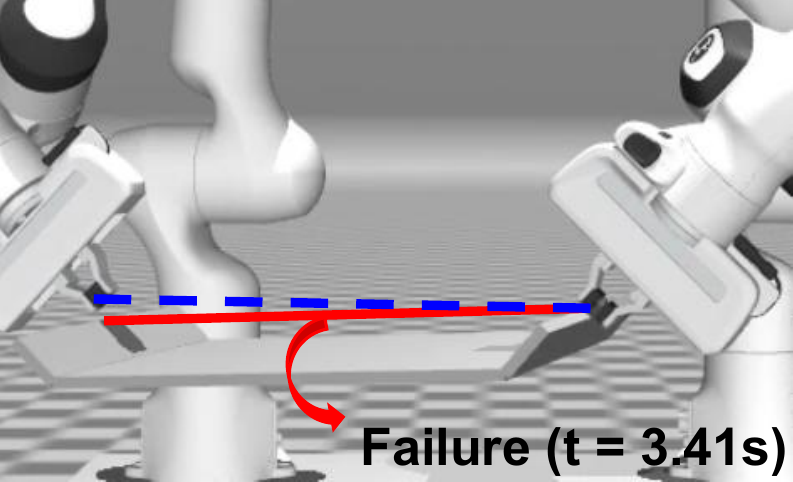}%
    \label{fig_2b}%
}\hfill
\subfloat[]{%
    \includegraphics[width=0.455\linewidth]{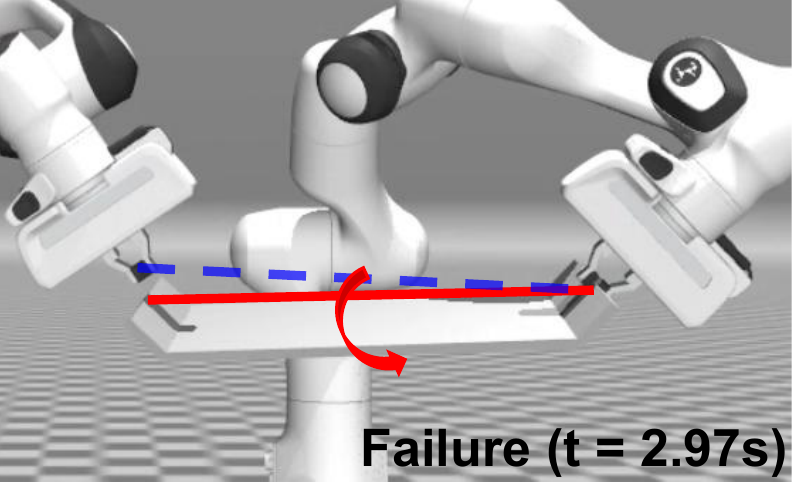}%
    \label{fig_2c}%
}
\caption{Snapshots of the equality-constraint validation experiment: (a) MC-MPPI successfully tracks the goal; (b) Latent MPPI fails at $t = 3.41$~s; (c) Vanilla MPPI fails at $t = 2.97$~s.}
\label{fig_2}

\subfloat[]{%
    \includegraphics[width=\linewidth]{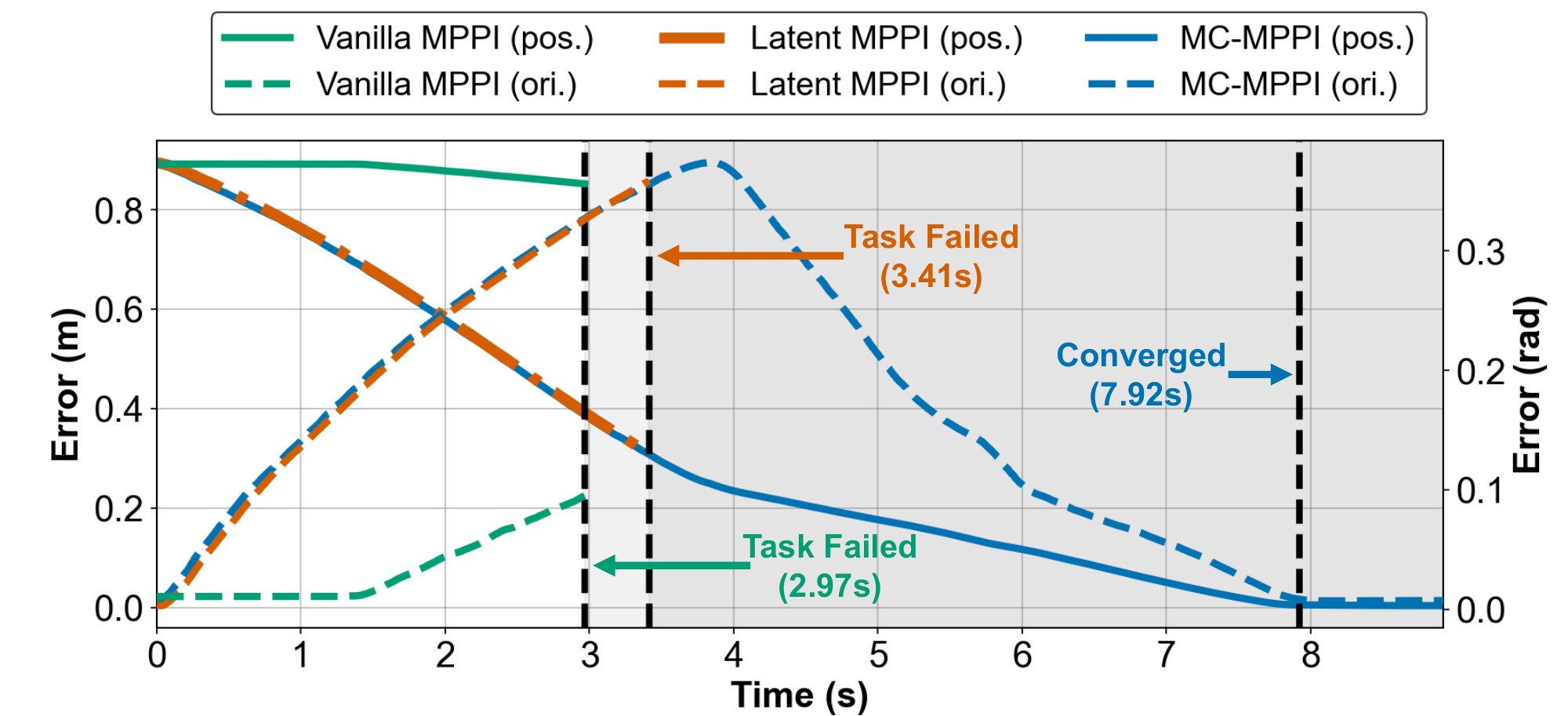}%
    \label{fig_3a}%
}\\
\subfloat[]{%
    \includegraphics[width=\linewidth]{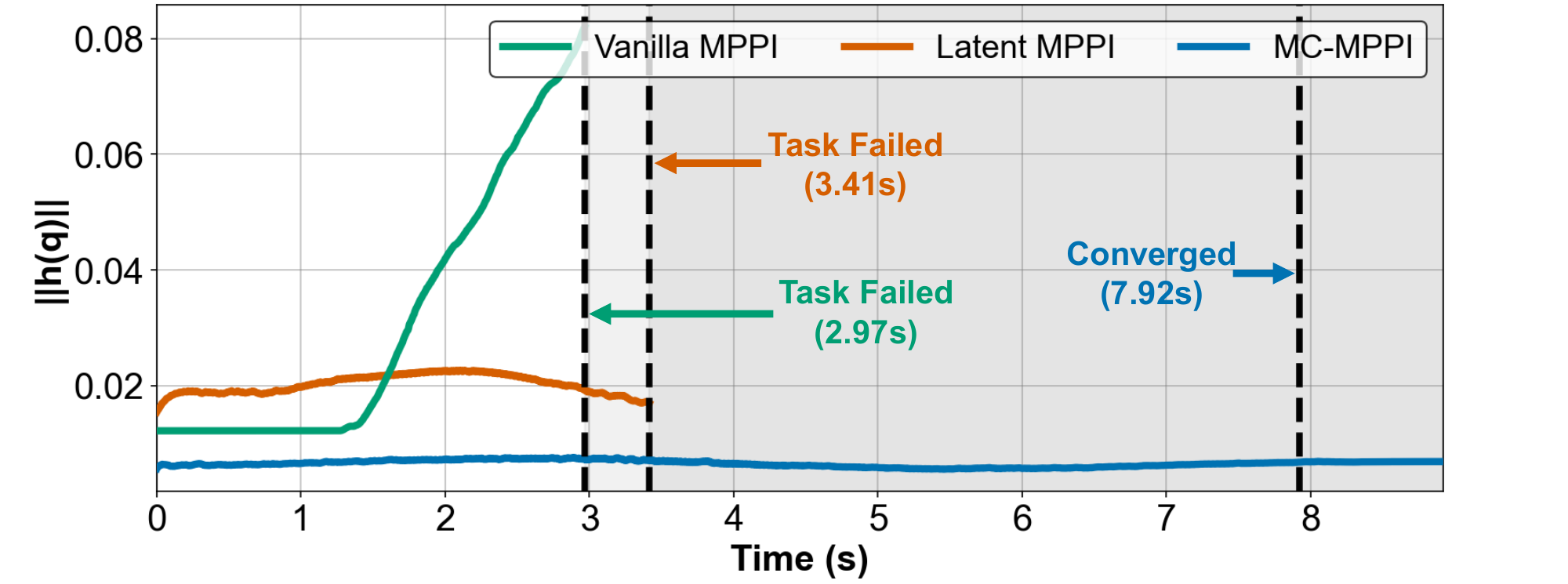}%
    \label{fig_3b}%
}
\caption{Quantitative results of the equality-constraint validation experiment: (a) tracking error and (b) equality-constraint violation.}
\label{fig_3}
\end{figure}

\subsection{\texorpdfstring{Validation of Equality-Constraint Regulation}{Validation of Equality-Constraint Regulation}}
\label{sec_42}

This experiment investigates the necessity of the QP-based execution stage by comparing the proposed MC-MPPI framework against two baselines: \textit{Latent MPPI}, an ablated version of MC-MPPI that retains the VAE-based planning stage of Section \mbox{\ref{sec_31}} but omits the execution stage of Section \mbox{\ref{sec_32}}, and \textit{Vanilla MPPI}, which performs standard MPPI of Section \mbox{\ref{sec_2}} directly in the joint configuration space with $\boldsymbol{u} = \dot{\boldsymbol{q}}$, where the equality constraint is treated through a large penalty cost on $\|\boldsymbol{h}(\boldsymbol{q})\|$. The tray center is commanded to move from $(0.4, 0.4, 0.9)$~m to $(0.4, -0.4, 0.5)$~m, both with identity orientation. The standard deviation of the control perturbation $\delta\boldsymbol{u}^{(k)}$ is uniformly set to $0.005$ across all evaluated methods.

As shown in Figs.~\mbox{\ref{fig_2}} and~\mbox{\ref{fig_3}}, MC-MPPI successfully converges to the target at $7.92$~s while effectively satisfying the equality constraint $\boldsymbol{h}(\boldsymbol{q})$ throughout the motion. The framework achieves this by maintaining the violation norm below the $0.01$ threshold---yielding an average violation of $0.0066 \pm 0.0007$---which reliably preserves both the closed-chain ($\boldsymbol{h}_{\mathrm{cc}}$) and tray flatness ($\boldsymbol{h}_{\mathrm{flat}}$) terms.

In contrast, both Vanilla MPPI and Latent MPPI fail before reaching the target, but their failure modes differ markedly. Vanilla MPPI fails earliest at $2.97$~s, exhibiting an average violation of $0.0314$ with a peak of $0.0820$ while barely reducing the tracking error---reflecting the fundamental difficulty of finding feasible candidates in the ambient configuration space, where the soft penalty cannot constrain the chaotic exploration. Latent MPPI, by contrast, drives the tracking error down smoothly from the very first seconds, indicating that the VAE-learned latent space does enable meaningful exploration near the constraint manifold. However, despite this manifold-aware exploration, the residual manifold mismatch from the learned decoder---with an average violation of $0.0199$ and a peak of $0.0226$---accumulates over time and ultimately causes the bimanual grasp to break at $3.41$~s.

Together, these results validate the two-stage design of MC-MPPI: the VAE-based planning stage successfully generates near-feasible trajectories that lie close to the constraint manifold---producing the smooth tracking behavior also observed in Latent MPPI---while the QP-based execution stage regulates the residual manifold mismatch that Latent MPPI cannot resolve, thereby maintaining the constraint residual within the prescribed tolerance during execution.

\begin{figure*}[t]
\centering
\includegraphics[width=\linewidth]{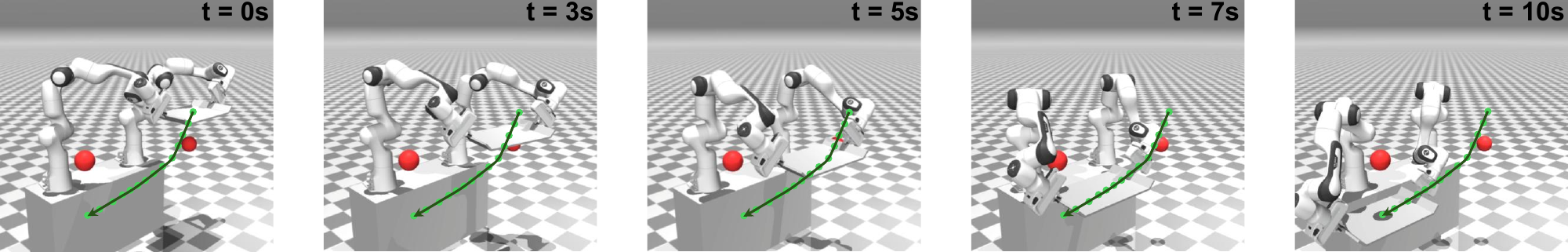}
\caption{
Representative snapshots of MC-MPPI with \eqref{eq_14}
in the static-obstacle experiment, where the two red spheres denote
the obstacles.
}
\label{fig_4}
\end{figure*}

\subsection{Obstacle Avoidance under Static Environments}
\label{sec_43}

\begin{figure}[t!]
\centering
\subfloat[]{%
    \includegraphics[width=\linewidth]{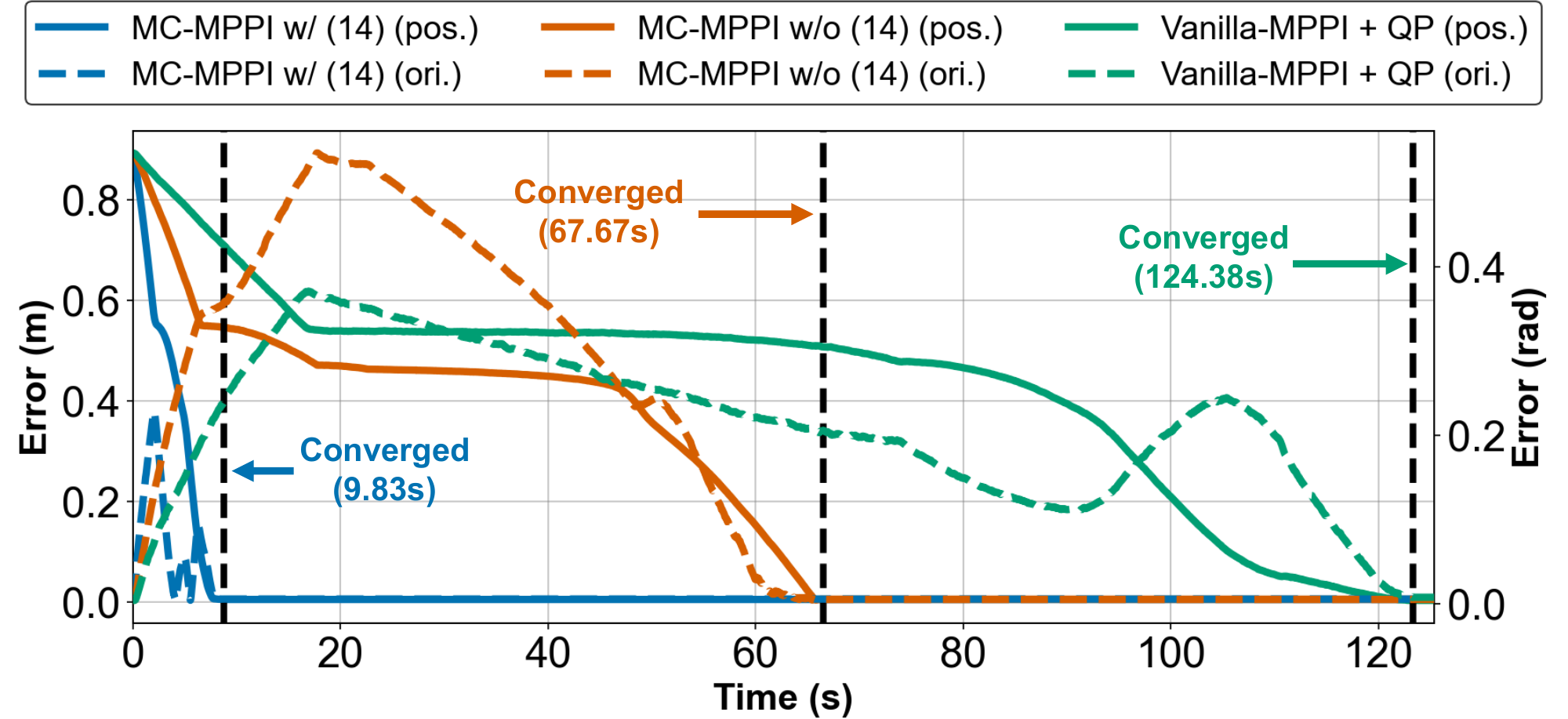}%
    \label{fig_5a}%
}\\
\subfloat[]{%
    \includegraphics[width=\linewidth]{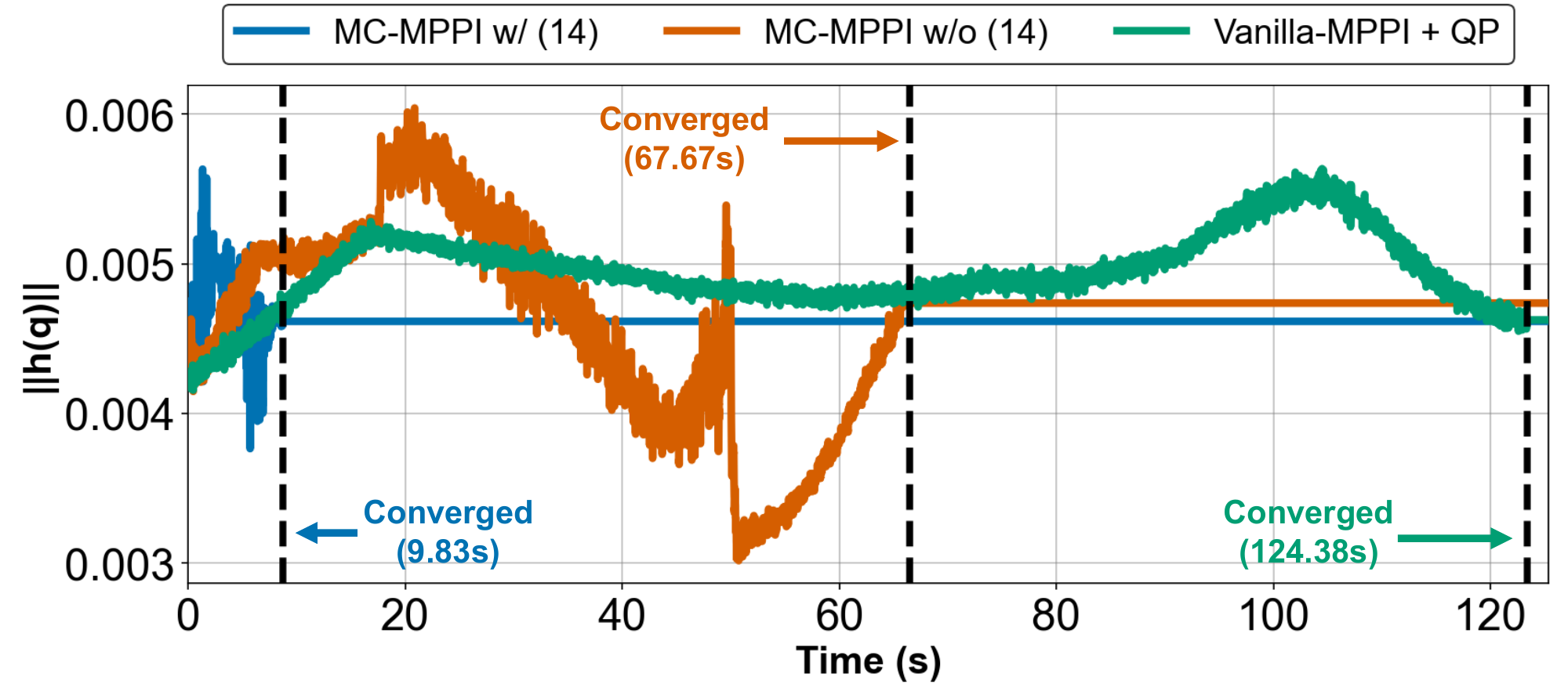}%
    \label{fig_5b}%
}
\caption{
Representative trial of the static-obstacle experiment:
(a) tracking error and (b) equality-constraint residual.
Vanilla MPPI~$+$~QP uses $w_h=10^{4}$.
}
\label{fig_5}
\end{figure}

This experiment evaluates static-obstacle avoidance under the
manifold-based equality constraint and compares MC-MPPI with the
constant-innovation sampling rule in \mbox{\eqref{eq_14}}, MC-MPPI
without \mbox{\eqref{eq_14}}, and Vanilla MPPI~$+$~QP.
Here, MC-MPPI without \mbox{\eqref{eq_14}} retains the same latent
representation, rollout cost, sample count, planning horizon, and
execution QP, but replaces the constant innovation with independent
per-step perturbations.
Vanilla MPPI~$+$~QP uses the same execution QP but performs planning
directly in the ambient joint space and adds the equality penalty
$w_h S_{\mathrm{eq}}$ to its rollout cost.

The tray center is commanded to move from $(0.4, 0.4, 0.9)$~m to
$(0.4, -0.4, 0.5)$~m with identity orientation.
Two spherical obstacles of radius $5$~cm, centered at
$(0.4, 0.4, 0.6)$~m and $(0.4, -0.4, 0.7)$~m, block the direct path.
The latent perturbation standard deviation is $0.01$ for both MC-MPPI
variants.
All methods are evaluated over $N_{\mathrm{run}}=10$ independent
random seeds with a timeout of $150$~s per trial.

Table~\mbox{\ref{tab:exp2_static}} summarizes the repeated-run results, while Figs.~\mbox{\ref{fig_4}} and~\mbox{\ref{fig_5}} show representative trajectories under the same task setting and random seed. Both MC-MPPI variants completed all ten trials. MC-MPPI with \mbox{\eqref{eq_14}} converged in $10.115\pm0.265$~s, approximately seven times faster than MC-MPPI without \mbox{\eqref{eq_14}}, which required $72.843\pm5.457$~s. The decoder generates near-feasible rollouts before the tracking and collision costs are evaluated, while the constant innovation in \mbox{\eqref{eq_14}} produces smooth and directionally sustained exploration along the learned manifold.

However, Vanilla MPPI~$+$~QP is highly sensitive to the equality-penalty weight $w_h$. With $w_h=10^{3}$, off-manifold rollouts are insufficiently suppressed, and the subsequent equality correction can invalidate the collision geometry evaluated during planning, causing all ten trials to fail by collision. With $w_h=10^{4}$, all ten trials succeed, but the representative trajectory in Fig.~\mbox{\ref{fig_5a}} remains nearly stationary near the obstacles for approximately $70$~s, reflecting the scarcity of samples that are simultaneously near-feasible, collision-free, and task-progressing and resulting in a mean convergence time of $120.267\pm2.409$~s. With $w_h=10^{5}$, this sampling scarcity becomes more severe because the equality and obstacle costs assign large costs to most rollouts, and all ten trials time out near the obstacles. These results highlight the strong sensitivity of Vanilla MPPI~$+$~QP to the choice of $w_h$.

Overall, these results show that near-feasible sampling in MC-MPPI substantially improves planning quality and obstacle-avoidance progress without requiring task-dependent equality-penalty tuning. Meanwhile, Fig.~\mbox{\ref{fig_5b}} shows similarly small equality-constraint residuals for all three successful configurations, indicating that the shared execution-stage QP consistently regulates the remaining residual regardless of the planning representation. The comparison therefore supports the proposed two-stage design: MC-MPPI improves the quality of the planning reference, while the execution stage provides reliable local constraint regulation.

As shown in Table~\mbox{\ref{tab:exp2_static}}, the planning time increased only slightly from $6.119\pm0.214$~ms for Vanilla MPPI~$+$~QP to $6.4228\pm0.0785$~ms for MC-MPPI with \mbox{\eqref{eq_14}}. Batched decoder inference required $0.9511\pm0.0085$~ms, but this is not a direct additive overhead because Vanilla MPPI instead evaluates the equality residual and penalty throughout the rollout batch. Thus, the net increase was only about $0.3$~ms, and MC-MPPI remained within the $10$~ms budget for $100$~Hz planning.

\begin{table}[t]
\centering
\caption{Repeated-run results for the static-obstacle experiment
over ten independent random seeds.}
\label{tab:exp2_static}
\resizebox{\linewidth}{!}{%
\begin{tabular}{lccc}
\toprule
Method
& \shortstack{Success\\ / Total}
& \shortstack{Convergence\\Time (s)}
& \shortstack{Planning\\Time (ms)} \\
\midrule
Vanilla MPPI $+$ QP ($w_h=10^{3}$)
& $0/10$
& --
& $6.182 \pm 0.0755$ \\

Vanilla MPPI $+$ QP ($w_h=10^{4}$)
& $10/10$
& $120.267 \pm 2.409$
& $6.119 \pm 0.214$ \\

Vanilla MPPI $+$ QP ($w_h=10^{5}$)
& $0/10$
& --
& $6.187 \pm 0.0965$ \\

MC-MPPI w/o \eqref{eq_14}
& $10/10$
& $72.843 \pm 5.457$
& $6.3891 \pm 0.0613$ \\

MC-MPPI w/ \eqref{eq_14}
& $10/10$
& $10.115 \pm 0.265$
& $6.4228 \pm 0.0785$ \\
\bottomrule
\end{tabular}%
}
\end{table}

\begin{figure*}[t]
\centering
\includegraphics[width=1.0\linewidth]{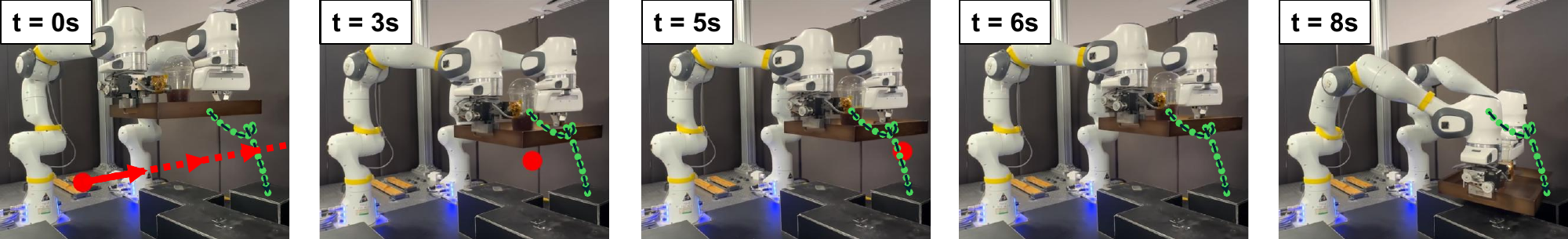}
\caption{Snapshots of the dynamic obstacle avoidance experiment for a representative trial on the real dual-arm system. A virtual moving sphere is introduced by feeding its trajectory directly into the controller, and the system successfully executes real-time evasion maneuvers while maintaining the manifold-based equality constraints.}

\label{fig_6}
\end{figure*}

\begin{figure}[!t]
\subfloat[]{%
    \includegraphics[width=\linewidth]{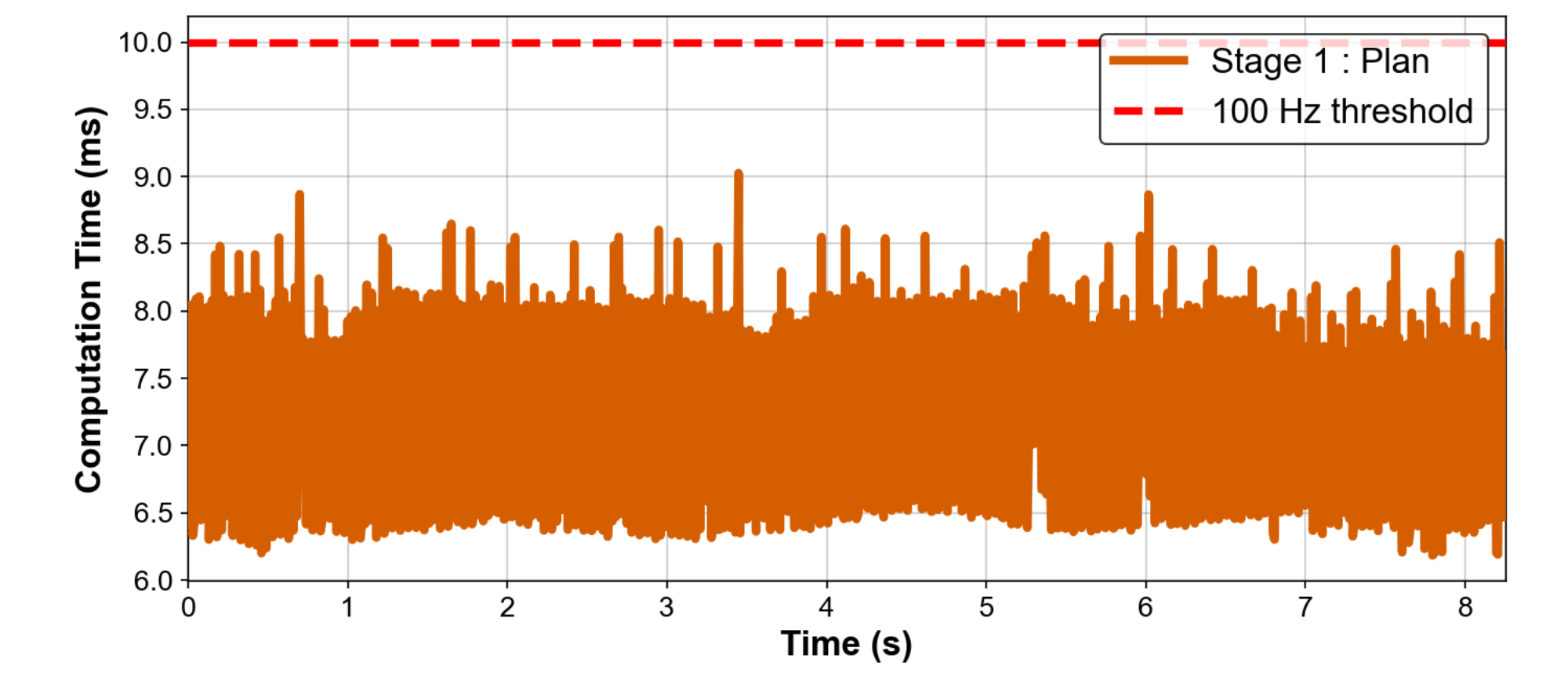}%
    \label{fig_7a}%
}\\
\subfloat[]{%
    \includegraphics[width=\linewidth]{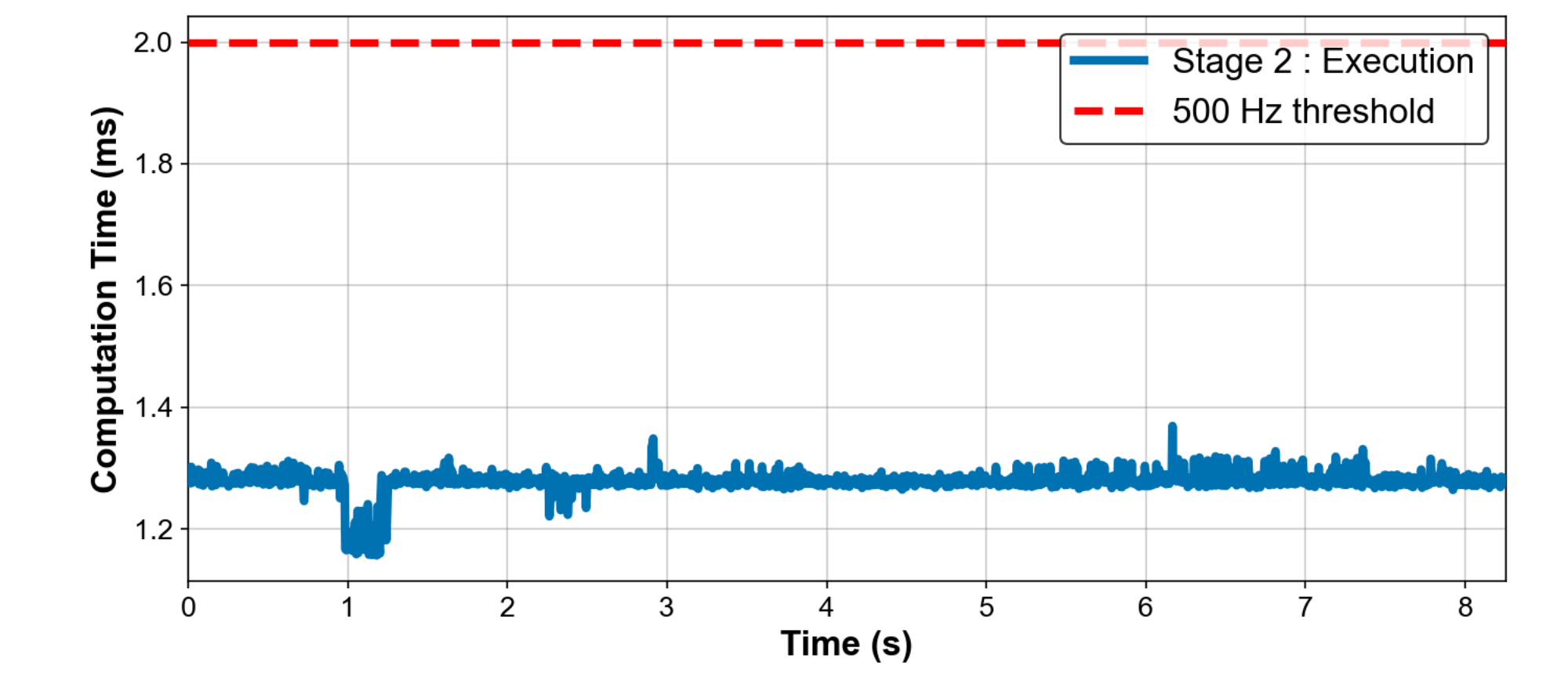}%
    \label{fig_7b}%
}
\caption{Computation time of the dynamic obstacle avoidance experiment for a representative trial: (a) planning stage cycle time, consistently within the 10~ms budget (100~Hz); (b) execution stage cycle time, consistently within the 2~ms budget (500~Hz).}
\label{fig_7}
\end{figure}

\subsection{Obstacle Avoidance under Dynamic Environments}
\label{sec_44}
This experiment evaluates the reactiveness of MC-MPPI on the real closed-chain dual-arm hardware under time-varying environments. The bimanual system is commanded to transport a tray between uniformly sampled poses, where both the initial and goal configurations are drawn from $x \in [0.4, 0.6]$~m, $y \in [-0.3, 0.3]$~m, $\text{yaw} \in [-\pi/6, \pi/6]$, with $z \in [0.8, 1.0]$~m and $z \in [0.4, 0.6]$~m, respectively. During each trial, a single spherical obstacle of radius $5$~cm traverses the workspace along either the $x$- or $y$-axis at a randomly selected speed of $0.1$ or $0.2$~m/s. A total of 40 trials are conducted under these randomized conditions.

Among the 40 trials, 38 are successfully completed, yielding a $95\%$ success rate. As illustrated in Fig.~\mbox{\ref{fig_6}}, the system avoids the moving obstacle while maintaining an average constraint violation of $0.0067 \pm 0.0010$. Latent-space replanning allows the controller to react to the observed obstacle motion without an explicit motion-prediction model. In the representative trial shown in Fig.~\mbox{\ref{fig_7}}, the planning and execution processes have cycle times of $7.098 \pm 0.729$~ms (maximum $9.025$~ms) and $1.277 \pm 0.020$~ms (maximum $1.369$~ms), respectively, meeting the $100$~Hz and $500$~Hz timing requirements. The two unsuccessful trials occur when the obstacle moves along the $x$-axis at the higher tested speed. In both unsuccessful trials, the equality residual remained below the prescribed threshold and the grasp was preserved, but the system collided with the obstacle. The observed motions indicate that the required lateral displacement exceeded the available closed-chain workspace within the reaction horizon. This limitation may be mitigated by obstacle-motion prediction, a larger reachable workspace, or a longer planning horizon.

\section{Conclusion}
\label{sec_5}
This paper presented Manifold-Constrained MPPI (MC-MPPI), a real-time sampling-based control framework for robotic systems subject to manifold-based equality constraints. MC-MPPI combines VAE-based latent-space planning with QP-based execution, retaining MPPI's derivative-free sampling while regulating the residual equality-constraint error at execution. Simulation and hardware experiments on a 14-DoF closed-chain dual-arm system demonstrated operation at $100$~Hz planning and $500$~Hz execution while regulating the constraints in static and dynamic obstacle scenarios.

Although the architecture can be applied to other robotic systems whose feasible configurations form a kinematic equality manifold, the current implementation has three main limitations. First, the learned latent space is tied to a specific manifold and requires retraining or conditioning when the robot kinematics, grasp transform, or task geometry changes. Second, collision avoidance remains a planning cost and is not guaranteed by the execution QP. Third, the present analysis does not establish formal robustness to measurement noise, modeling error, or communication delay. Future work will address these limitations through conditioned generative models for parameterized manifolds, explicit safety mechanisms for infeasible QP updates, and tractable integration of inequality constraints into the execution stage.

\section*{Declarations}

\subsection*{Conflict of Interest}
The authors declare that there is no competing financial interest or personal relationship that could have appeared to influence the work reported in this paper.

\subsection*{Authors' Contributions}
Seulchan Lee contributed to the implementation, conducted the experiments, and participated in writing the manuscript. Sanghyun Kim led the research direction and contributed to the writing of the manuscript.

\subsection*{Funding}
This research was supported by the National Research Foundation of Korea (NRF) grant funded by the Korea government (MSIT) (No. RS-2024-00411007), the National Research Council of Science \& Technology (NST) grant funded by the Korea government (MSIT) (No. GTL25042-000), and the Korea Basic Science Institute (National Research Facilities and Equipment Center) grant funded by the Ministry of Science and ICT (No. RS-2025-00564593).

\bibliographystyle{IEEEtran}
\bibliography{IEEEabrv, references}

\end{document}